\documentclass[letterpaper,10 pt,conference]{IEEEtran}
\IEEEoverridecommandlockouts
\usepackage{cite}
\usepackage{amsmath,amssymb,amsfonts}
\usepackage{algorithmic}
\usepackage{graphics} 
\usepackage{epsfig} 
\usepackage{academicons}

\usepackage{graphicx}
\usepackage[hidelinks]{hyperref}
\usepackage{textcomp}
\usepackage{xcolor}
\usepackage{amsmath}
\usepackage{mathptmx} 

\usepackage{amssymb}  

\usepackage{svg}
\usepackage{array}
\usepackage{graphicx}
\usepackage{svg}
\usepackage{comment}

\begin{document}

\title{\LARGE \bf Loopy Movements: Emergence of Rotation in a Multicellular Robot \\

\thanks{$^{1}$The authors are with the Department of Mechanical, Materials, and Aerospace Engineering, West Virginia University, Morgantown, WV 26505, USA}
\thanks{$^{2}$Trevor Smith is the corresponding author 
{\tt\small trs0024@mix.wvu.edu}}%

\thanks{This study was supported by the National Science Foundation Graduate Research Fellowship Award \#2136524 and NSF Award \#2422270}
}

\author{ $^*$Trevor Smith$^{1,2}$,  Yu Gu$^{1}$ \\
\textit{West Virginia University}\\
Morgantown, USA 
}

\maketitle

\begin{abstract}
Unlike most human-engineered systems, many biological systems rely on emergent behaviors from low-level interactions, enabling greater diversity and superior adaptation to complex, dynamic environments. This study explores emergent decentralized rotation in the Loopy multicellular robot, composed of homogeneous, physically linked, 1-degree-of-freedom cells. Inspired by biological systems like sunflowers, Loopy uses simple local interactions—diffusion, reaction, and active transport of simulated chemicals, called morphogens—without centralized control or knowledge of its global morphology. Through these interactions, the robot self-organizes to achieve coordinated rotational motion and forms lobes—local protrusions created by clusters of motor cells. This study investigates how these interactions drive Loopy’s rotation, the impact of its morphology, and its resilience to actuator failures. Our findings reveal two distinct behaviors: 1) inner valleys between lobes rotate faster than the outer peaks, contrasting with rigid body dynamics, and 2) cells rotate in the opposite direction of the overall morphology. The experiments show that while Loopy’s morphology does not affect its angular velocity relative to its cells, larger lobes increase cellular rotation and decrease morphology rotation relative to the environment. Even with up to one-third of its actuators disabled and significant morphological changes, Loopy maintains its rotational abilities, highlighting the potential of decentralized, bio-inspired strategies for resilient and adaptable robotic systems.

\end{abstract}

\section{Introduction}
Organisms across nature demonstrate diverse and fascinating motion mechanisms, often without centralized control. Sunflowers, for example, rotate to bask in sunlight through the collective emergent behavior of their cells, driven by simple interactions such as ion active transport to regulate hydraulic pressure and the production and diffusion of the growth hormone Auxin in response to light \cite{VANDENBRINK201420,atamian2016circadian,briggs2016sunflowers}. This work seeks to extend similar emergent motion to Loopy—a multi-cellular robot—by investigating how decentralized, homogeneous, and interconnected agents can self-organize to achieve coordinated movement, specifically focusing on rotational motion in this study. Here, emergence refers to self-organized behaviors that arise through local interactions among the agents without centralized planning or the use of global information \cite{smith2023swarm,slavkov2018Morphogenesis, Chatty2011Emergence}. 

\begin{figure}
    \centering
    \includegraphics[width=0.48 \textwidth]{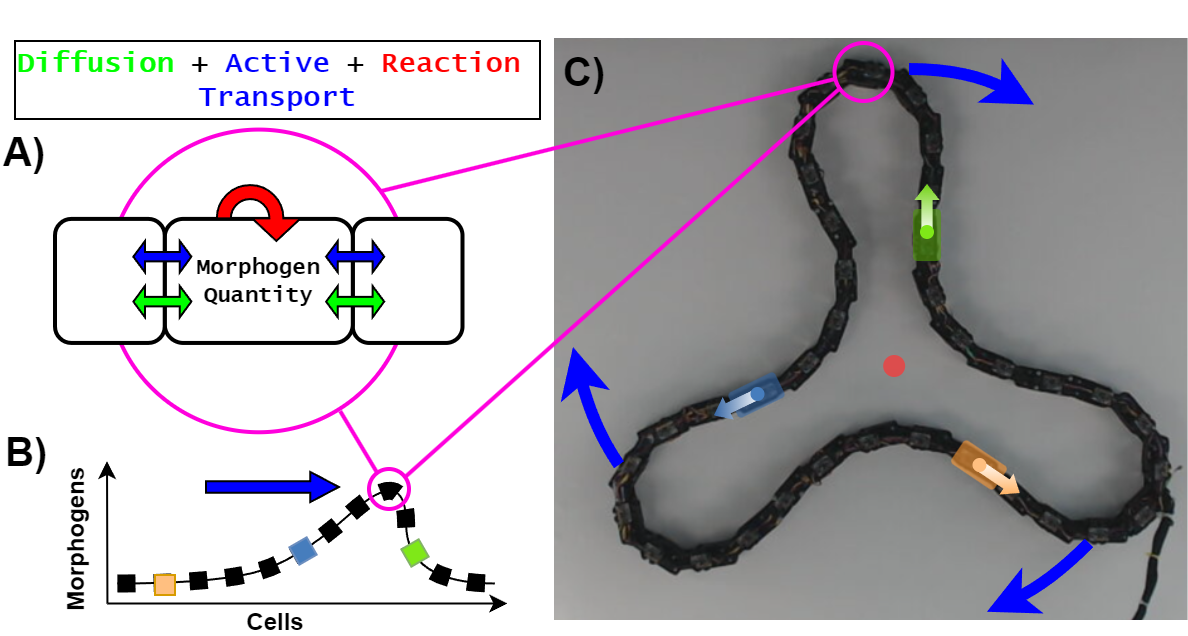}
    \caption{ A) Simulated cellular interactions, including diffusion, reaction, and active transport, enable the decentralized emergence and propagation of diverse chemical distributions. B) These chemical patterns propagate through Loopy’s cells, determining the desired joint angles of its actuators. C) This leads to the emergence of Loopy’s rotational movement about its centroid (red dot) and lobed morphology, where the cells rotate in the opposite direction to the overall morphology. }
    \label{visual_abstract}
\end{figure}

In the case of Loopy, as shown in Fig. \ref{visual_abstract}, each motor functions as a cell, forming a closed chain that communicates through simulated chemical reactions and diffusion (Fig. \ref{visual_abstract}.A) to create stable morphologies of lobed protrusions (Fig. \ref{visual_abstract}.C) \cite{smith2023swarm}. In this work, morphology refers to Loopy's overall global shape, which remains consistent regardless of which cells constitute different parts of the shape. Importantly, Loopy's cells are homogeneous in function, with no specialized roles or hierarchical distinctions, operating solely on local information without any awareness of Loopy's overall structure or shape.

Since Loopy was not purposely designed for coordinated movement, it faces unique and difficult challenges in achieving emergent motion. Although it resembles snake-like robots in its continuous chain structure, Loopy's homogeneous cells lack specialization initially, meaning it has no designated head or tail for orientation \cite{smith2023swarm}. Additionally, Loopy does not have directional friction nor the ability to lift itself out of the plane to guide motion \cite{gray1946mechanism, rollinson2016pipe,write2012snake, Ijspeert2007cpgsnake}. As a result, the approaches utilized by traditional snake robots that rely on these features \cite{gray1946mechanism, rollinson2016pipe,write2012snake, Ijspeert2007cpgsnake} are not applicable. Thus, new knowledge of how bulk motion can emerge from decentralized interactions among homogeneous, physically interlinked components is needed to empower Loopy with the capacity for movement. Achieving this could inspire the design of future robots composed of simple, redundant agents capable of adapting to dynamic environments and withstanding partial system failures. One potential application is the dynamic marking of the boundary of a large, moving hazardous or contaminated area, such as a chemical spill, adjusting in real-time to component failures and changing environmental conditions.

Traditional robotic motion relies on top-down approaches, where centralized systems control all actuators to follow pre-determined trajectories based on complete models of the robot's morphology, dynamics, and environment \cite{gu2018robot, gu2023stonemine, ohi2018bramblbee}. Even with advancements in machine learning, these systems still depend on centralized policies learned through extensive training \cite{atmeh2014mlatlas, deisenroth2012toward, Gangapurwala2022mllegged, grcic2021dense}. In contrast, bio-inspired robotics, which often model biological systems, may simulate neural circuits like central pattern generators (CPGs) to coordinate motion \cite{li2017neural,goldsmith2020neurodynamic,ijspeert2007swimming,zaier2008cpg}. However, even this approach requires central optimization, and while CPGs partially decentralize locomotion to groups of motors, it remains fundamentally a top-down design utilizing the given trajectories of biological counterparts.

Swarm robotics has demonstrated how decentralized coordination among agents using simple local interactions can lead to emergent bulk motion \cite{reynolds1987flocks, rubenstein2014programmable}. However, most swarm systems consist of physically disconnected agents interacting virtually. Meanwhile, modular and reconfigurable robotic swarms dynamically elect to physically connect or disconnect based on the current task \cite{baldassarre2007sbots,ozkan2021self}. While they may exhibit some decentralized behaviors \cite{baldassarre2007sbots,ozkan2021self}, these systems often rely on centralized control for configuration and coordination. Loopy differs fundamentally as a single, interlinked robot without knowledge of its morphology, in which each component is physically constrained to exert constant mechanical interactions with its neighbors, in addition to its decentralized virtual interactions.

This research seeks to address this gap in swarm robotics by investigating decentralized, physically connected agents to create emergent motion through local interactions. In doing so, we ask: 1) How do Loopy's rotational motions emerge from local interactions? 2) What role does Loopy's morphology play in its movement? 3) How resilient is this approach to individual component failures? Though answering these questions, this work contributes to the literature in the following ways:

\begin{enumerate}
    \item Establishing a link between simple cellular interactions, inspired by plant behavior, and the emergence of rotation in the multicellular Loopy robot.
    \item Experimentally demonstrating the influence of Loopy's morphology on its rotational behavior.
    \item Demonstrating Loopy's resilience to incremental component failures through experiments. 
\end{enumerate}

The rest of this paper is outlined as follows. Section \ref{methodology} details the methodology utilized in this work. Section \ref{exp} describes the experimental setup, results, and discussion of the experiments. Finally, Section \ref{future} concludes and expresses the future work of this study.

\section{Methodology}
\label{methodology}
\subsection{Problem Statement}

The objective of this study is to explore decentralized strategies for emergent rotation in the Loopy robot through simple cellular interactions. Loopy consists of a large number of homogeneous, 1 degree of freedom (DoF) cells that are physically linked. Each cell, constructed from a rotary servo, functions as an independent agent and communicates only with its immediate neighbors. Additionally, each cell lacks awareness of the robot's overall morphology or position.

While the sunflower’s motion is driven by external signals (sunlight) \cite{VANDENBRINK201420,atamian2016circadian,briggs2016sunflowers}, this study deliberately excludes external environmental influences. Loopy lacks extroceptive sensors, and focusing solely on internal interactions simplifies the system, ensuring that emergent motion arises from within. By constraining the environment to be obstacle-free, sensory-free, and flat, the study emphasizes internal decentralized coordination, minimizing reliance on external guidance. This approach allows us to investigate how rotation can emerge naturally, driven solely by local interactions, without predefined motion trajectories.

\subsection{Reaction, Diffusion, and Active Transport}
Similar to \cite{smith2023swarm} in this study, each motor in the Loopy platform is treated as an independent cell that performs simulated chemical reactions inside the cell and communicates with its neighbors through the diffusion of these simulated chemicals or morphogens as shown in Fig. \ref{visual_abstract}.A. Specifically, the Fitzhugh-Nagumo activator-inhibitor reaction \eqref{Qactivator}-\eqref{Qinhibitor} and a purely diffusive morphogen \eqref{Qpassive} are utilized to emerge sinusoidal morphogen distributions \cite{fitzhugh1955mathematical}.

\begin{equation}
    \dot{q}_{act} = \gamma_{act} \nabla^2 q_{act} +  \left(q_{act} - q_{act}^3 - q_{inh} + \alpha \right)
    \label{Qactivator}
\end{equation}

 \begin{equation}
    \dot{q}_{inh} = \gamma_{inh} \nabla^2 q_{inh} + \beta(q_{act} - q_{inh})
    \label{Qinhibitor}
\end{equation}

\begin{equation}
        \dot{q}_{pas} = \gamma_{pas} \nabla^2 q_{pas}
        \label{Qpassive}
\end{equation}

Where $q$ is the morphogen quantity, $\gamma$ is the diffusion rates, $\alpha$ is the stimulation rate, and $\beta$ is the inhibition rate \cite{fitzhugh1955mathematical}. The morphogen concentrations within each motor cell then dictate its goal joint angle ($\theta$) with \eqref{M2ang}. 

\begin{equation}
        \theta = q_{pas} + q_{act}
        \label{M2ang}
\end{equation}

\begin{figure}
    \centering
    \includegraphics[width=0.3 \textwidth]{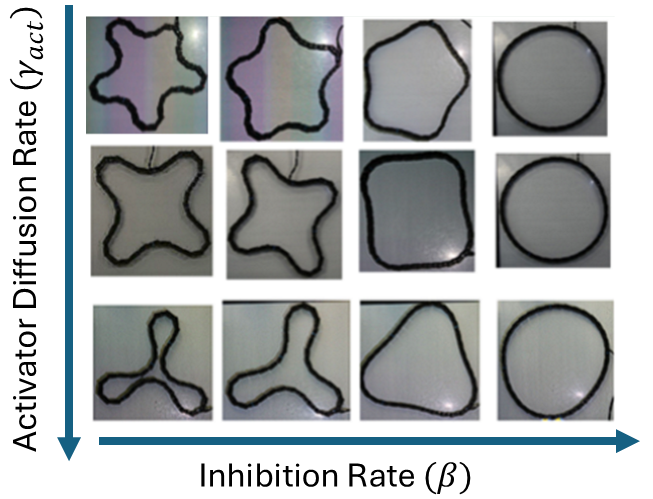}
    \caption{The morphology space of the Loopy robot. Increasing the inhibition rate ($\beta$) reduces lobe size, while increasing the activator diffusion rate ($\gamma_{act}$) decreases the number of lobes formed.}
    \label{morphspace}
\end{figure}

The balance between diffusion and reaction parameters directly influences Loopy's global shape, or morphology. For example, increasing the inhibition rate $\beta$ limits activator accumulation, causing Loopy’s lobes—local protrusions formed by clusters of motor cells—to shrink. Conversely, increasing the activator diffusion rate $\gamma_{act}$ allows the activator to spread across more cells, which reduces the number of lobes formed (Fig. \ref{morphspace}) \cite{smith2023swarm}.

In addition to passive chemical interactions (reactions and diffusion), active transport is crucial for sustaining motion \cite{cheng2015design}. Without it, passive interactions will eventually settle into equilibrium, causing movement to stop \cite{cheng2015design}. Active transport disrupts this equilibrium by moving chemicals against their concentration gradients \cite{cheng2015design}, maintaining the conditions necessary for continuous coordinated motion. A simple active transport model is shown in \eqref{wave}, where $\boldsymbol{Q}$ represents the morphogen quantities, and $v$ is the transportation rate \cite{chen2014dynamic, bacstuug2012molecular}.

\begin{equation}
    \dot{\boldsymbol{Q}} = v \nabla \boldsymbol{Q}
    \label{wave}
\end{equation}

Upon closer inspection, \eqref{wave} is a first-order wave equation where $v$ represents the wave speed. This equation allows any morphogen distribution to propagate at a constant speed around Loopy's cellular ring, as shown in Fig. \ref{visual_abstract}.B. Since the morphogen concentration in each cell determines its joint angle \eqref{M2ang}, the entire morphology—independent of which cells form specific parts—propagates around the chain, inducing rotation at an angular velocity $\omega_{mc}$ relative to its cells. This angular velocity is determined by \eqref{m_rot_c} where $N$ is the number of cells and $s$ is the length of each cell.

\begin{equation}
    \omega_{mc} = v \frac{2\pi}{N \cdot s}
    \label{m_rot_c}
\end{equation}

To solve and decentralize the continuous partial differential equations \eqref{Qactivator}-\eqref{Qpassive}, they are discretized both temporally and spatially, respectively, with each motor cell of size $s$ serving as a control volume \cite{smith2023swarm}. The active transport gradient and second-order diffusion gradient are approximated using first-order and second-order central difference methods, respectively. Thus, \eqref{Qactivator}-\eqref{Qpassive}, with the inclusion of active transport \eqref{wave}, become \eqref{Qad}-\eqref{Qpd}, respectively, for the $m^{th}$ cell in the loop. These equations are expressed as a sum of active transport, diffusion, and reaction terms, with each term enclosed in parentheses. The system is then propagated through time ($t$) using \eqref{time_update}, where $\Delta t$ is the time-step.

    \begin{equation}
        \begin{aligned}
        \begin{split}
         \dot{q}_{act_{m,t}} & = v \left(\frac{q_{act_{m+1,t}} - q_{act_{m-1,t}}}{2s}\right) \\
         & +  \gamma_{act} \left(\frac{q_{act_{m-1,t}} - 2q_{act_{m,t}} + q_{act_{m+1,t}}}{2s}\right) \\
         & +  \left(q_{act_{m,t}} - q_{act_{m,t}}^3 - q_{inh_{m,t}} + \alpha \right)
        \end{split}
        \label{Qad}
        \end{aligned}
    \end{equation}

    \begin{equation}
        \begin{aligned}
        \begin{split}
            \dot{q}_{inh_{m,t}} & = v \left(\frac{q_{inh_{m+1,t}} - q_{inh_{m-1,t}}}{2s}\right) \\
            & + \gamma_{inh} \left(\frac{q_{inh_{m-1,t}} - 2q_{inh_{m,t}} + q_{inh_{m+1,t}}}{2s}\right) \\
            & + \beta \left(q_{act_{m,t}} - q_{inh_{m,t}}\right)
        \end{split}
        \label{Qid}
        \end{aligned}
    \end{equation}

    \begin{equation}
        \begin{aligned}
        \begin{split}
            \dot{q}_{pas_{m,t}} & = v \left(\frac{q_{pass_{m+1,t}} - q_{pass_{m-1,t}}}{2s}\right) \\
             & +  \gamma_{pas} \left(\frac{q_{pas_{m-1,t}} - 2q_{pas_{m,t}} + q_{pas_{m+1,t}}}{2s}\right)
        \end{split}
        \end{aligned}
        \label{Qpd}
    \end{equation}

    \begin{equation}
        \boldsymbol{Q}_{m,t+1} =  \dot{\boldsymbol{Q}}_{m,t}\Delta t + \boldsymbol{Q}_{m,t} 
        \label{time_update}
    \end{equation}

From \eqref{Qad}-\eqref{Qpd}, it is clear that the $m^{th}$ motor cell relies solely on the morphogen state of its own current time step ($t$) and the adjacent cells, $m-1$ and $m+1$. This decentralizes the process, as each cell operates using only local information from its immediate neighbors.

\section{Experiments, Results, and Discussion}
\label{exp}

\subsection{Experimental Setup}
To assess Loopy's rotational capabilities, four experiments were conducted with step changes in the following parameters: 1) wave speed, 2) lobe size, 3) number of lobes, and 4) number of failed actuators. Vicon markers tracked each motor cell's position, which was converted into polar coordinates relative to Loopy's centroid. Numerical differentiation was then applied to calculate angular velocities. In addition, to reduce noise from numerical differentiation, the angular velocities were smoothed using a convolution with a centered rectangular kernel, covering 5\% of the step window.

\begin{figure}
    \centering
    \includegraphics[width=0.3 \textwidth]{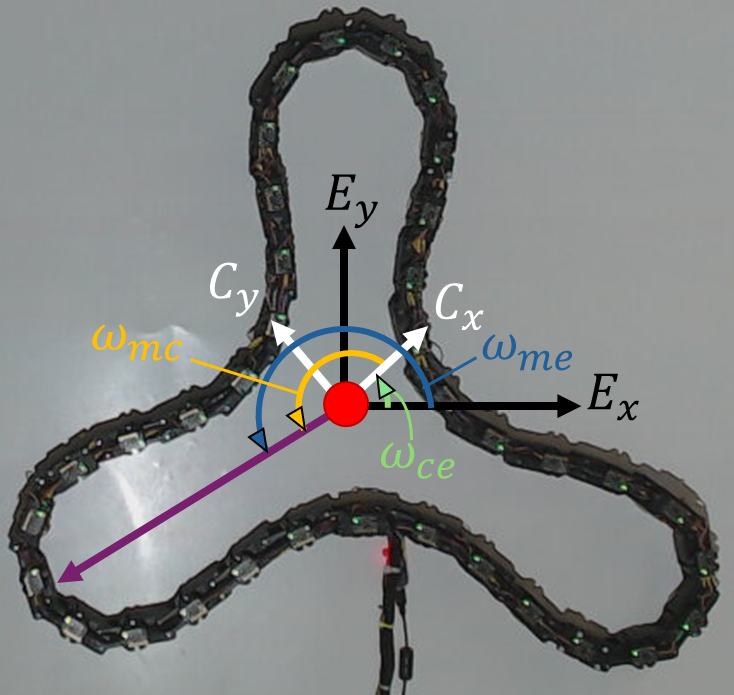}
    \caption{ Two coordinate frames are utilized, both centered on Loopy's centroid (red circle): $E$ (black), fixed to the environment, and $C$ (white), fixed to Loopy's cells, with $C_x$ pointing to the first cell. The rotation of $C$ relative to $E$ describes the angular velocity of Loopy's cells in the environment ($\omega_{ce}$, green). The rotation of Loopy's morphology, indicated by the motion of its lobe peaks (purple), is described by $\omega_{me}$ (blue). Lastly, $\omega_{mc}$ (yellow) represents the rate at which the morphology shifts along Loopy's closed cellular chain.}
    \label{loopy_frames}
\end{figure}

The three recorded angular velocities are illustrated in Fig. \ref{loopy_frames}, with two reference frames: $E$ (fixed to the environment) and $C$ (fixed to Loopy's cells), where $C_x$ points to the first cell. The angular velocity between these frames represents the rotation of the cells relative to the environment ($\omega_{ce}$). The angular velocity of the morphology relative to the environment ($\omega_{me}$) captures the rotation of Loopy's lobes in the environment, while $\omega_{mc}$ measures the rate at which the lobed morphology shifts along the cellular chain, driven by morphogen propagation through active transport. Furthermore, from Fig. \ref{loopy_frames}, the morphology's rotation in the environment is the sum of its rotation relative to its cells and the cell's rotation in the environment \eqref{frame_match}.

\begin{equation}
    \omega_{me} = \omega_{mc} + \omega_{ce}
    \label{frame_match}
\end{equation}

\subsection{Demonstration of Rotation}
\begin{figure}
    \centering
    \includegraphics[width=0.35 \textwidth]{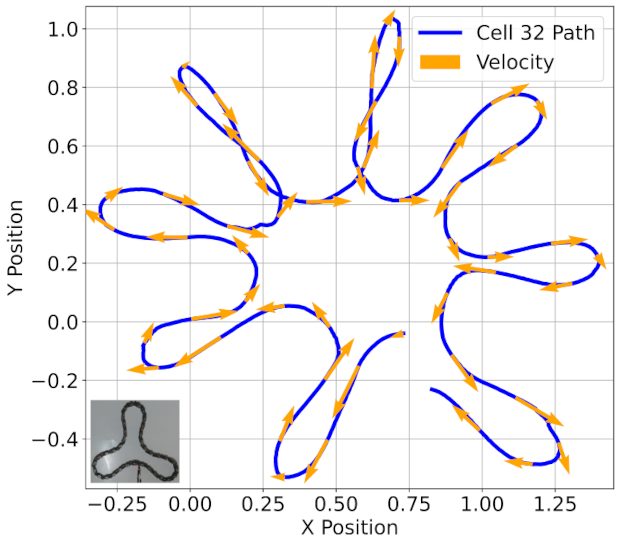}
    \caption{A representative path (blue) and corresponding velocities (orange) of one of Loopy's cells as it completes a full environment cycle in a three-lobed morphology. The tangential velocity is highest in the valleys between the lobes and approaches zero at the tips of the lobes.}
    \label{path}
\end{figure}

\begin{figure}
    \centering
    \includegraphics[width=0.47 \textwidth]{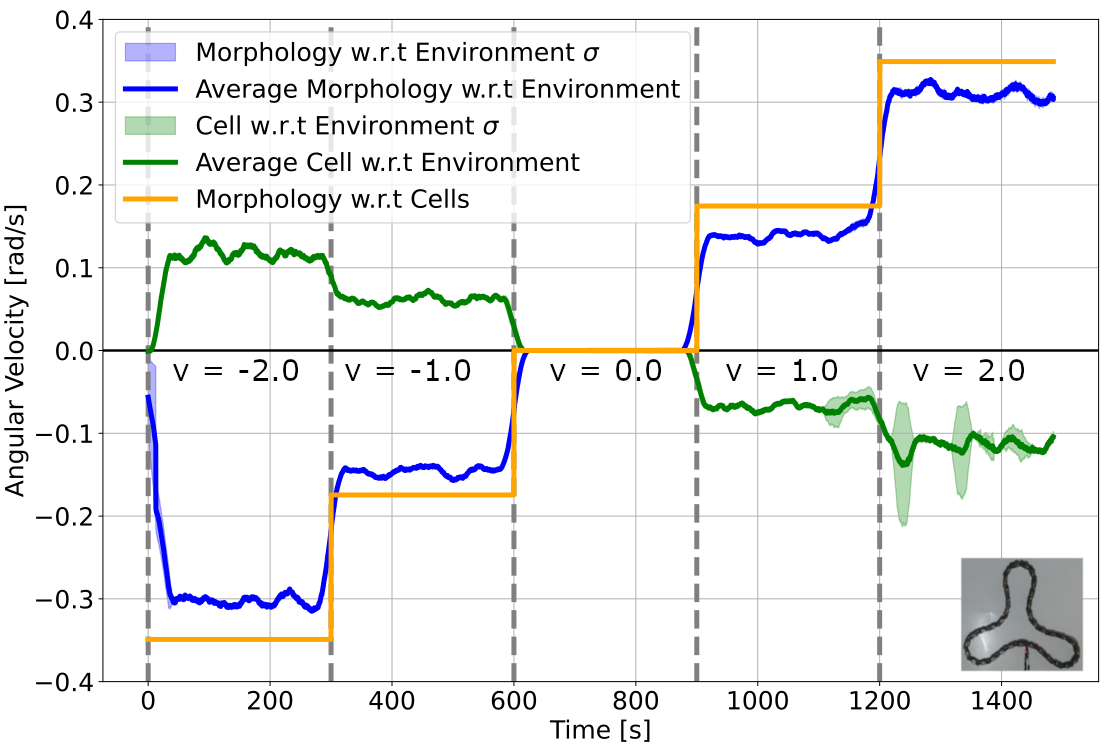}
    \caption{A three-lobed Loopy subjected to step changes in wave speed $v$ (indicated by grey dashed sections). The mean angular velocities of Loopy's cells, along with their first standard deviation ($\sigma$), are displayed. As wave speed magnitude increased, all angular velocities increased. The sign of $v$ determined the direction of rotation, with Loopy's cells (green) rotating opposite to its morphology (blue and yellow).}
    \label{demo_exp}
\end{figure}

\begin{figure}
    \centering
    \includegraphics[width=0.48 \textwidth]{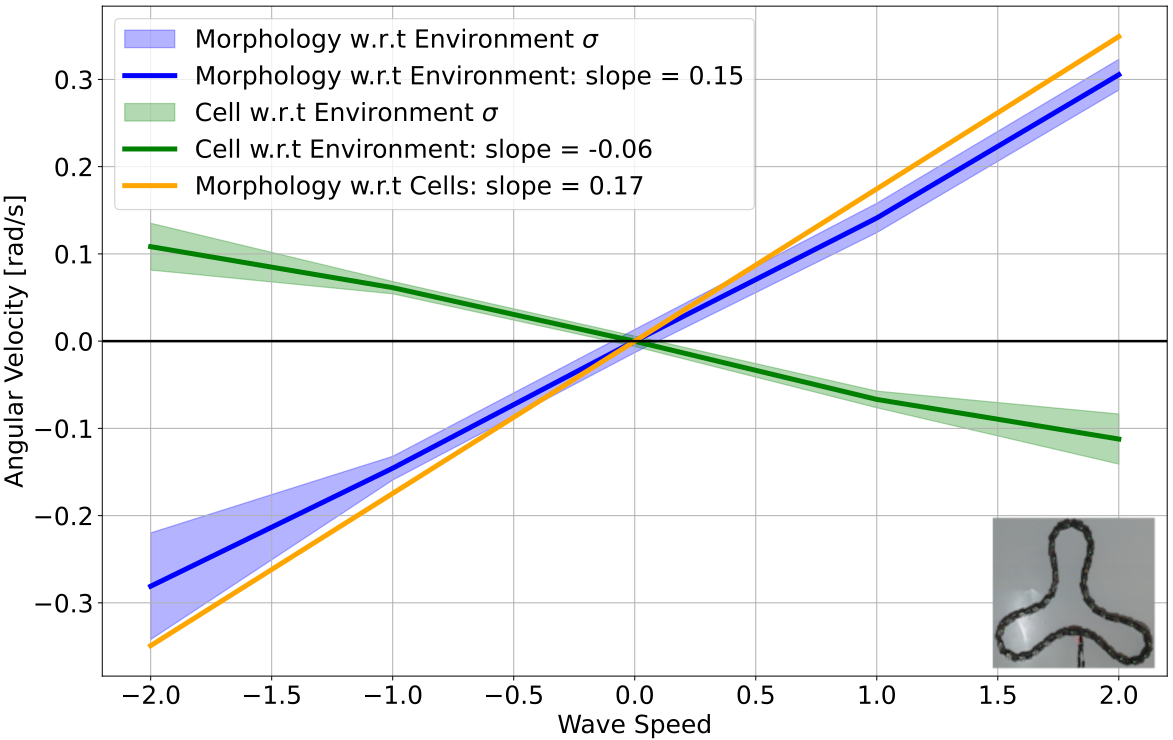}
    \caption{The average angular velocities of Loopy's cells, along with the first standard deviation ($\sigma$), as a function of the wave speed parameter $v$. The wave speed parameter had a linear effect on all three rotational velocities within the tested range, and the linear slope is displayed on the figure legend.}
    \label{vel_vs_speed}
\end{figure}

In this experiment, a three-lobed Loopy with moderate lobe size ($\gamma_{inh} = 100, \gamma_{act} = 1.0$, $\beta = 225$, and $\alpha = 0.001$) was subjected to step changes in wave speed ($v$) from $-2.0$ to $2.0$ in increments of 1.0, with each step lasting five minutes. A representative path of one full environmental rotation from a single Loopy cell is shown in Fig. \ref{path}. The mean angular velocities across all cells, along with their standard deviations, are displayed in Fig. \ref{demo_exp}. The blue curve represents the morphology's angular velocity with respect to the environment ($\omega_{me}$), the green curve shows the cells' angular velocity relative to the environment ($\omega_{ce}$), and the yellow line shows the analytical calculation of the morphology's angular velocity with respect to its cells ($\omega_{mc}$). Fig. \ref{vel_vs_speed} presents the three angular velocities as a function of wave speed ($v$) along with their linear fits.

Inspecting Fig. \ref{path} shows that the highest velocities occurred along the radial components of Loopy’s lobes, predominantly in directions that do not directly contribute to rotation. In contrast, the inner valleys between the lobes exhibited the greatest velocity tangential to the centroid, driving the cellular rotation ($\omega_{ce}$). Interestingly, the velocity was minimal at the tips of the lobes, suggesting Loopy's rotation is driven by twisting the inter-lobe valleys forward while the lobe tips remain stationary. This is in direct contrast to rigid body rotation, where the shorter the radial distance, the slower the tangential velocity. Although Loopy is not a single rigid body and does not obey this constraint, its morphology is maintained as it passes along its cells, which makes Loopy appear as a rigid body at the macroscopic level. 

As shown in Fig. \ref{demo_exp} and Fig. \ref{vel_vs_speed}, increasing the magnitude of the wave speed parameter ($v$) led to proportional increases in the magnitudes of all three angular velocities ($\omega_{mc}$, $\omega_{me}$, and $\omega_{ce}$), even though $\omega_{ce}$ (green) rotates in the opposite direction and becomes more negative. This is expected, as increasing the magnitude of $v$ accelerates morphogen propagation along the cellular chain, directly increasing the morphology's rotation about its cells. This, in turn increases the Loopy's environmental rotations. Additionally, the direction of rotation is determined by the sign of $v$, with all velocities affected linearly within the tested range (Fig. \ref{vel_vs_speed}), exhibiting rapid, smooth (Fig. \ref{demo_exp}), and predictable transitions.

An additional observation from Fig. \ref{demo_exp} and Fig. \ref{vel_vs_speed} is that increasing the wave speed ($v$) amplifies the difference between $\omega_{mc}$ and $\omega_{me}$. According to \eqref{frame_match}, the morphology's rotation relative to the environment ($\omega_{me}$) is the sum of its rotation relative to the cells ($\omega_{mc}$) and the cells' rotation relative to the environment ($\omega_{ce}$). As $\omega_{ce}$ increases, the gap between $\omega_{mc}$ and $\omega_{me}$ widens accordingly.

However, Fig. \ref{vel_vs_speed} shows that the sum of $\omega_{mc}$ and $\omega_{ce}$ does not perfectly equal $\omega_{me}$, likely due to the lobes reducing Loopy's effective circumference. This discrepancy arises because \eqref{m_rot_c} assumes a small wave height, but Loopy's large lobes distort its overall geometry, leading to a mismatch.

The next observation of Fig. \ref{demo_exp} and \ref{vel_vs_speed} is that Loopy's cells rotated clockwise in the environment, while Loopy's morphology rotated counterclockwise. This is interesting because if Loopy is observed as a whole at the macroscopic scale, it would appear to be rotating counterclockwise. However, if Loopy is observed microscopically at the cellular level, each cell moves clockwise in the environment, which directly contrasts with the macroscopic observation. 

\subsection{Lobe Size Experiment}

\begin{figure}
    \centering
    \includegraphics[width=0.45 \textwidth]{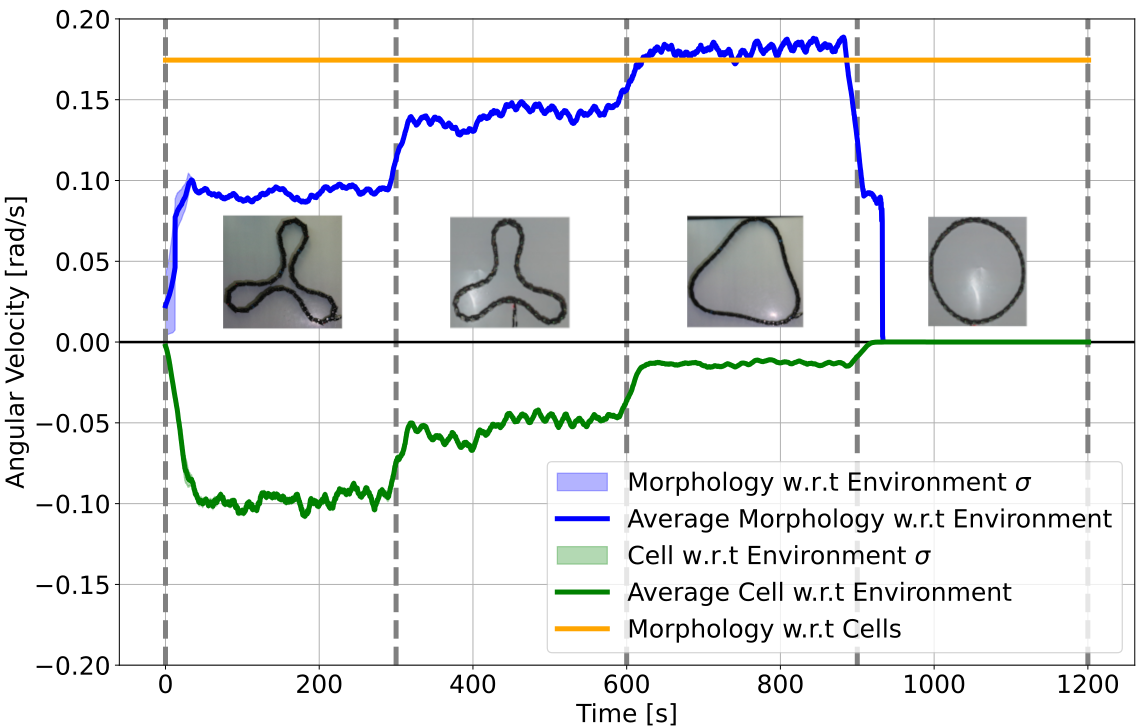}
    \caption{Loopy's mean angular velocities and first standard deviation ($\sigma$) as lobe size decreases. The morphology's angular velocity relative to the cells (yellow) remains constant, while its velocity relative to the environment (blue) increases, and the cells' angular velocity (green) decreases. Rotation stops when lobes disappear, with minimal variation in cellular angular velocity. }
    \label{size_exp}
\end{figure}

This experiment evaluates how changes in lobe size affect Loopy's rotational behavior. A three-lobed Loopy, with a constant wave speed ($\gamma_{inh} = 100$, $\gamma_{act} = 1.0$, $\alpha = 0.001$, and $v=1.0$), was subjected to five-minute step changes in $\beta$. Values of $\beta$ were chosen as [100, 225, 300, 500] to represent four distinct morphologies: 1) maximum lobe size, 2) moderate lobe size, 3) convex shape, and 4) circle, as shown in Fig. \ref{morphspace}. The mean angular velocities ($\omega_{mc}$, $\omega_{me}$, and $\omega_{ce}$) across all cells, along with their standard deviations, are presented in Fig. \ref{size_exp}.

The first observation from Fig. \ref{size_exp} is that the circular lobe shape stopped all angular velocities because the quantity of morphogens in the cells became identical, eliminating the morphogen gradient and halting wave propagation \eqref{wave}. Secondly, the angular velocity's standard deviation between cells is minimal in Fig. \ref{size_exp}, indicating little variation and high coordination among the cells. Thirdly, the angular velocity of Loopy's morphology relative to its cells ($\omega_{mc}$) remained constant regardless of lobe size, as morphogen wave propagation is independent of the wave shape. Fourth, as lobe size decreased, the cells' angular velocity ($\omega_{ce}$) also decreased, while the morphology's angular velocity relative to the environment ($\omega_{me}$) increased. This occurs because $\omega_{mc}$ is constant and $\omega_{ce}$ is decreasing. Thus, from \eqref{frame_match}, $\omega_{me}$ must increase.This effect allows Loopy to redistribute its morphology's angular velocity relative to its cells ($\omega_{mc}$), which is set by the wave speed $v$, between its macroscopic rotation in the environment ($\omega_{me}$) and microscopic cellular rotation ($\omega_{ce}$) by changing its lobe size through the $\beta$ parameter.

\subsection{Number of Lobes Experiment}
\begin{figure}
    \centering
    \includegraphics[width=0.49 \textwidth]{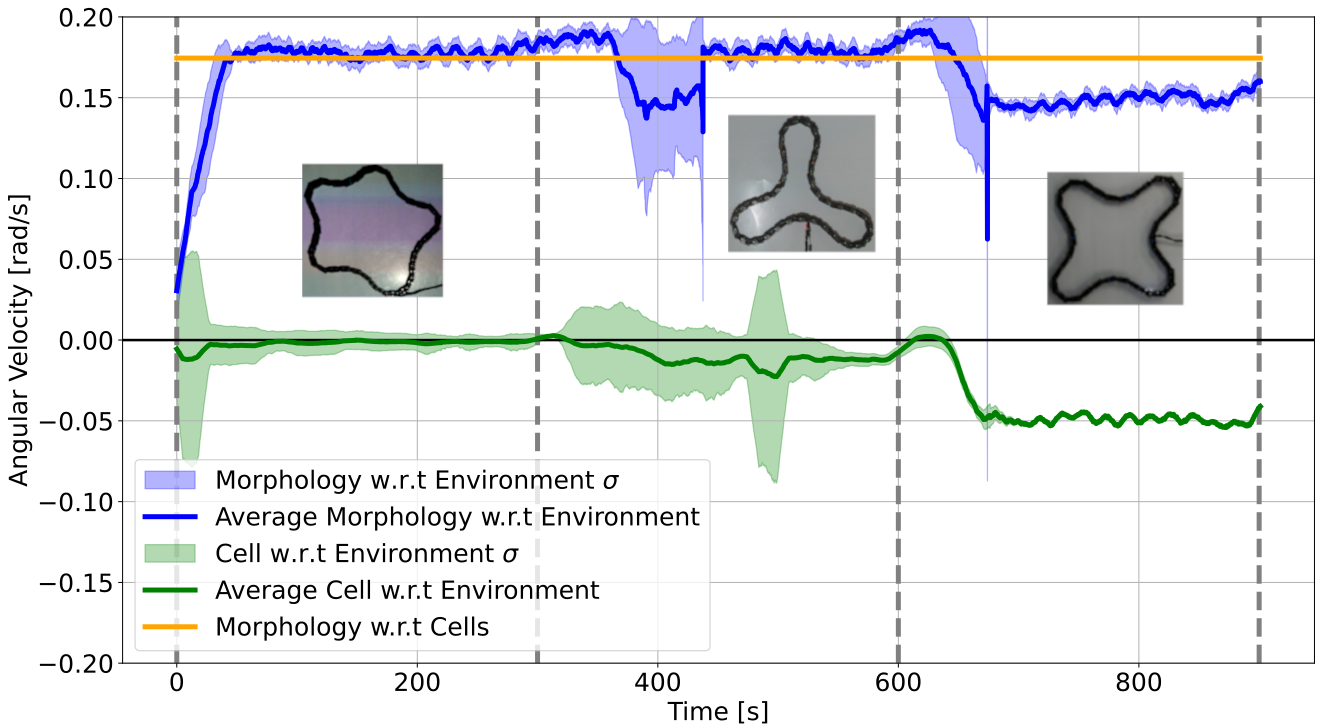}
    \caption{Loopy's mean angular velocities and first standard deviation ($\sigma$) as the number of lobes decreases. The morphology's angular velocity relative to the cells (yellow) stays constant, while its velocity relative to the environment (blue) decreases slightly. The cells' angular velocity (green) is zero with five lobes but increases in magnitude as lobes decrease. Standard deviation increased during morphology transitions.}
    \label{num_exp}
\end{figure}

This experiment evaluated how changes in the number of lobes affected Loopy's rotational abilities. Loopy ($\gamma_{inh} = 100$, $\beta = 225$, $\alpha = 0.001$, and $v = 1.0$) underwent five-minute step changes in its activator diffusion rate ($\gamma_{act}$), with values of [0.4, 0.9, 1.3] corresponding to 5, 4, and 3 lobe morphologies, as shown in Fig. \ref{morphspace}. The mean angular velocities for $\omega_{mc}$, $\omega_{me}$, and $\omega_{ce}$ across all cells, along with their first standard deviation, are presented in Fig. \ref{num_exp}.

Fig. \ref{num_exp} shows that as the number of lobes decreased, the cell's angular velocity ($\omega_{ce}$) increased, with the five-lobed morphology not rotating in the environment. This likely occurred because more lobes resulted in smaller lobes, which decreases $\omega_{ce}$ as observed in Section \ref{exp}.C. There is a coupling of the number of lobes and their size because Loopy is of finite length; thus as the number of lobes increases, there are fewer motors per lobe, resulting in each lobe being smaller. Additionally, the morphology's angular velocity relative to the cells ($\omega_{mc}$) remained constant and was similar to its angular velocity relative to the environment ($\omega_{me}$). This occurred for similar reasons to Section \ref{exp}.B, where $\omega_{mc}$ depends only on the wave speed parameter. In addition, $\omega_{me}$ was similar to environmental rotation due to the small cellular angular velocity ($\omega_{ce}$). Thus, from \eqref{frame_match}, the two morphology rotations must be comparable.

From our analytical expectations of \eqref{m_rot_c}, properties of the wave equation \eqref{wave}, Section \ref{exp}.C, and this experiment, it is evident that the rotation of Loopy's morphology with respect to its cells is independent of its morphology. However, the distribution of this rotation between cellular and morphology in the environment depends on the morphological shape. Nonetheless, a rotation mode remains feasible despite these notable changes in shape.

\subsection{Actuator Failure Experiment}

\begin{figure}
    \centering
    \includegraphics[width=0.45 \textwidth]{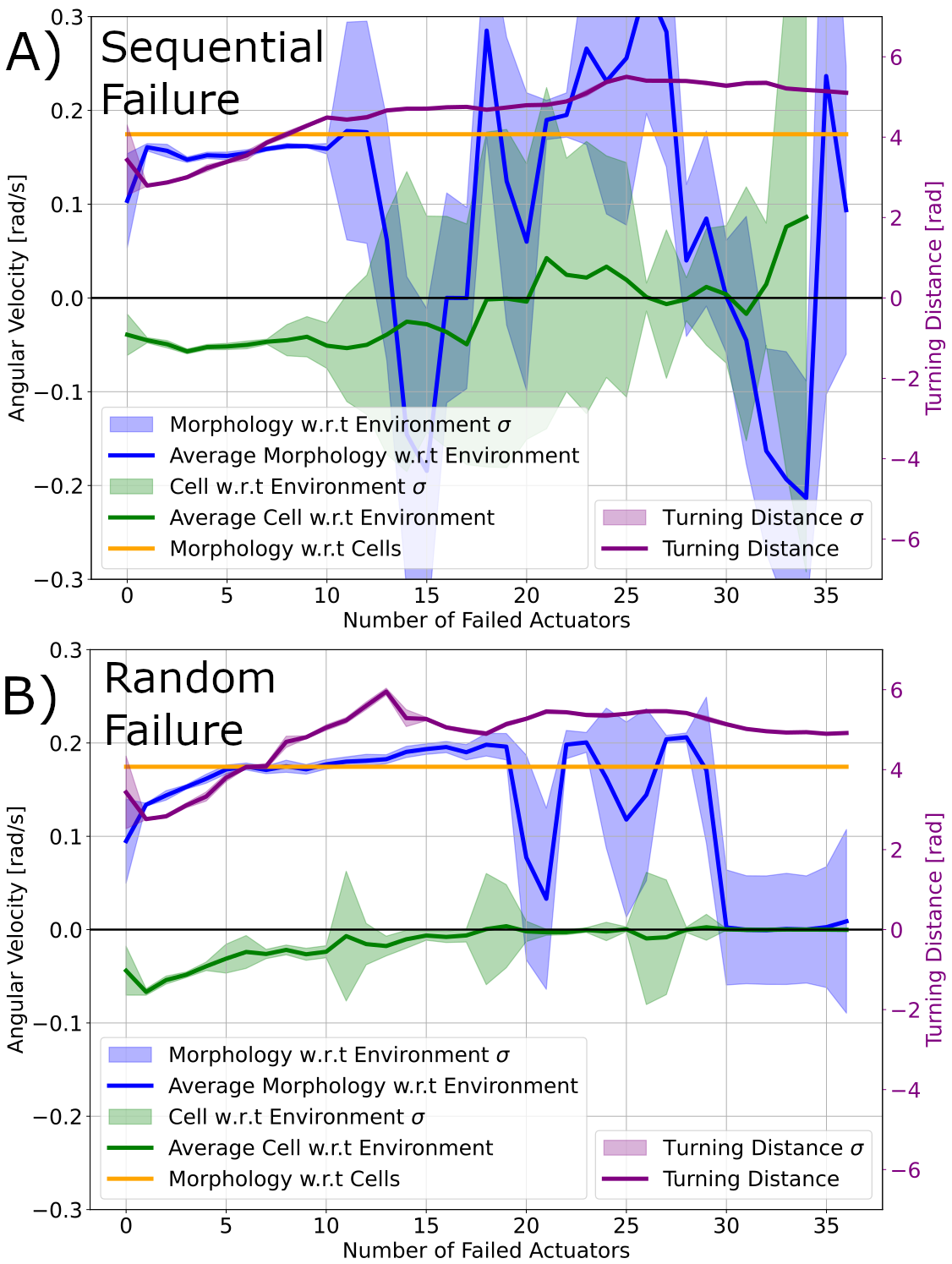}
    \caption{Loopy's mean angular velocities and first standard deviation ($\sigma$) during sequential (A) and random (B) actuator failures, along with the turning distance of its morphology. Loopy maintained rotation until about one-third of its motors failed. Sequential failures led to consistent rotation until a critical threshold, while random failures showed a gradual decline. Turning distance increased with motor failures, plateauing when half were inactive.}
    \label{failure}
\end{figure}

This experiment tested Loopy's resilience to incremental actuator failure. A three-lobed Loopy with moderate-sized lobes ($\gamma_{inh} = 100$, $\gamma_{act} = 1.0$, $\beta = 225$, $\alpha = 0.001$, and $v = 1.0$) underwent sequential and random failure modes. In sequential failure, motor torque was disabled one-by-one around the chain until all were disabled, while in random failure, motors were disabled in random order, simulating physical failures like gear teeth shearing. Angular velocities ($\omega_{mc}$, $\omega_{me}$, and $\omega_{ce}$) and the turning distance \cite{veltkamp2001shape}---a metric assessing shape degradation---were recorded (Fig. \ref{failure}).

The first observation from Fig. \ref{failure} is that Loopy continued rotating until approximately twelve motors (1/3 of its actuators) failed in both failure modes. In the sequential failure mode, cellular rotation ($\omega_{ce}$) remained consistent until a critical threshold, while in the random failure mode, rotation gradually degraded until it stopped. In both scenarios, torque-enabled cells pushed torque-disabled cells together, forming a rigid clump that allowed Loopy to keep rotating despite losing significant actuation. This explains why sequential failure had a threshold, as a continuous portion of active motors could still coordinate effectively, whereas random failure distributed inactive motors, leading to passive cancellation of active rotations when adjacent motors failed.

Next, the turning distance in Fig. \ref{failure} shows a gradual degradation of Loopy's morphology with each motor failure, plateauing once about half of the motors are disabled. At this point, the remaining motors lack the strength to move the disabled ones, halting all motion. In addition, Loopy's morphology becomes unrecognizable, leading to chaotic behavior of its rotations. Furthermore, despite enduring substantial damage of losing approximately a third of its motors, Loopy's morphology and cells continued rotating in the environment.

\section{Conclusions and Future Work}
\label{future}

This study investigated how decentralized, homogeneous, and interconnected agents can self-organize to achieve coordinated movement, using the Loopy multi-cellular robot as an example system. We addressed three key questions: First, how do Loopy's rotational motions emerge from local interactions? Loopy’s motions arise from diffusion, reaction, and active transport, without centralized control or knowledge of the overall shape. Two unique behaviors were observed: 1) inner valleys between lobes rotate faster than outer peaks, contrary to rigid body dynamics, and 2) cells rotate in the opposite direction of the overall morphology. Second, what role does Loopy's morphology play in its movement? While morphology did not affect angular velocity relative to the cells, larger lobes increased cellular rotation and reduced morphology rotation relative to the environment. Third, how resilient is this approach to component failures? Loopy continued rotating despite significant changes in morphology and losing one-third of its actuators. Although primitive, these findings highlight the potential of decentralized, bio-inspired strategies for creating adaptable, resilient, high-degree-of-freedom robotic systems that can adjust both behavior and morphology to complex, dynamic environments.

Future work will focus on physically augmenting Loopy to improve traction and increase rotational speed, as well as investigating the emergence of translational movements. Additionally, future studies will explore more complex terrains, incorporating fixed and movable obstacles to examine how Loopy interacts with its environment through deflection, deformation, and interacting with obstacles, potentially demonstrating emergent problem-solving abilities in navigating challenging terrains.

\section*{Acknowledgment}
We thank Thomas Smith of the University of Texas at Dallas for his insightful discussions on cellular biology.

\bibliographystyle{IEEEtran}
\bibliography{references}

\begin{thebibliography}{10}
\providecommand{\url}[1]{#1}
\csname url@samestyle\endcsname
\providecommand{\newblock}{\relax}
\providecommand{\bibinfo}[2]{#2}
\providecommand{\BIBentrySTDinterwordspacing}{\spaceskip=0pt\relax}
\providecommand{\BIBentryALTinterwordstretchfactor}{4}
\providecommand{\BIBentryALTinterwordspacing}{\spaceskip=\fontdimen2\font plus
\BIBentryALTinterwordstretchfactor\fontdimen3\font minus \fontdimen4\font\relax}
\providecommand{\BIBforeignlanguage}[2]{{%
\expandafter\ifx\csname l@#1\endcsname\relax
\typeout{** WARNING: IEEEtran.bst: No hyphenation pattern has been}%
\typeout{** loaded for the language `#1'. Using the pattern for}%
\typeout{** the default language instead.}%
\else
\language=\csname l@#1\endcsname
\fi
#2}}
\providecommand{\BIBdecl}{\relax}
\BIBdecl

\bibitem{VANDENBRINK201420}
\BIBentryALTinterwordspacing
J.~P. Vandenbrink, R.~I. Brown, S.~L. Harmer, and B.~K. Blackman, ``Turning heads: The biology of solar tracking in sunflower,'' \emph{Plant Science}, vol. 224, pp. 20--26, 2014. [Online]. Available: \url{https://www.sciencedirect.com/science/article/pii/S0168945214000818}
\BIBentrySTDinterwordspacing

\bibitem{atamian2016circadian}
H.~S. Atamian, N.~M. Creux, R.~I. Brown, A.~G. Garner, B.~K. Blackman, and S.~L. Harmer, ``Circadian regulation of sunflower heliotropism, floral orientation, and pollinator visits,'' \emph{Science}, vol. 353, no. 6299, pp. 587--590, 2016.

\bibitem{briggs2016sunflowers}
W.~R. Briggs, ``How do sunflowers follow the sun and to what end?'' \emph{Science}, vol. 353, no. 6299, pp. 541--542, 2016.

\bibitem{smith2023swarm}
T.~Smith, R.~M. Butts, N.~Adkins, and Y.~Gu, ``Swarm of one: Bottom-up emergence of stable robot bodies from identical cells,'' \emph{arXiv preprint arXiv:2306.12629}, 2023.

\bibitem{slavkov2018Morphogenesis}
\BIBentryALTinterwordspacing
I.~Slavkov, D.~Carrillo-Zapata, N.~Carranza, X.~Diego, F.~Jansson, J.~Kaandorp, S.~Hauert, and J.~Sharpe, ``Morphogenesis in robot swarms,'' \emph{Science Robotics}, vol.~3, no.~25, p. eaau9178, 2018. [Online]. Available: \url{https://www.science.org/doi/abs/10.1126/scirobotics.aau9178}
\BIBentrySTDinterwordspacing

\bibitem{Chatty2011Emergence}
A.~Chatty, I.~Kallel, P.~Gaussier, and A.~M. Alimi, ``Emergent complex behaviors for swarm robotic systems by local rules,'' in \emph{2011 IEEE Workshop on Robotic Intelligence In Informationally Structured Space}, 2011, pp. 69--76.

\bibitem{gray1946mechanism}
J.~Gray, ``The mechanism of locomotion in snakes,'' \emph{Journal of experimental biology}, vol.~23, no.~2, pp. 101--120, 1946.

\bibitem{rollinson2016pipe}
D.~Rollinson and H.~Choset, ``Pipe network locomotion with a snake robot,'' \emph{Journal of Field Robotics}, vol.~33, no.~3, pp. 322--336, 2016.

\bibitem{write2012snake}
C.~Wright, A.~Buchan, B.~Brown, J.~Geist, M.~Schwerin, D.~Rollinson, M.~Tesch, and H.~Choset, ``Design and architecture of the unified modular snake robot,'' in \emph{2012 IEEE International Conference on Robotics and Automation}, 2012, pp. 4347--4354.

\bibitem{Ijspeert2007cpgsnake}
A.~J. Ijspeert and A.~Crespi, ``Online trajectory generation in an amphibious snake robot using a lamprey-like central pattern generator model,'' in \emph{Proceedings 2007 IEEE International Conference on Robotics and Automation}, 2007, pp. 262--268.

\bibitem{gu2018robot}
Y.~Gu, J.~Strader, N.~Ohi, S.~Harper, K.~Lassak, C.~Yang, L.~Kogan, B.~Hu, M.~Gramlich, R.~Kavi \emph{et~al.}, ``Robot foraging: Autonomous sample return in a large outdoor environment,'' \emph{IEEE Robotics \& Automation Magazine}, vol.~25, no.~3, pp. 93--101, 2018.

\bibitem{gu2023stonemine}
C.~Tatsch, J.~A. Bredu, D.~Covell, I.~B. Tulu, and Y.~Gu, ``Rhino: An autonomous robot for mapping underground mine environments,'' in \emph{2023 IEEE/ASME International Conference on Advanced Intelligent Mechatronics (AIM)}, 2023, pp. 1166--1173.

\bibitem{ohi2018bramblbee}
N.~Ohi, K.~Lassak, R.~Watson, J.~Strader, Y.~Du, C.~Yang, G.~Hedrick, J.~Nguyen, S.~Harper, D.~Reynolds, C.~Kilic, J.~Hikes, S.~Mills, C.~Castle, B.~Buzzo, N.~Waterland, J.~Gross, Y.-L. Park, X.~Li, and Y.~Gu, ``Design of an autonomous precision pollination robot,'' in \emph{2018 IEEE/RSJ International Conference on Intelligent Robots and Systems (IROS)}, 2018, pp. 7711--7718.

\bibitem{atmeh2014mlatlas}
G.~M. Atmeh, I.~Ranatunga, D.~O. Popa, K.~Subbarao, F.~Lewis, and P.~Rowe, ``Implementation of an adaptive, model free, learning controller on the atlas robot,'' in \emph{2014 American Control Conference}, 2014, pp. 2887--2892.

\bibitem{deisenroth2012toward}
M.~P. Deisenroth, R.~Calandra, A.~Seyfarth, and J.~Peters, ``Toward fast policy search for learning legged locomotion,'' in \emph{2012 IEEE/RSJ International Conference on Intelligent Robots and Systems}.\hskip 1em plus 0.5em minus 0.4em\relax IEEE, 2012, pp. 1787--1792.

\bibitem{Gangapurwala2022mllegged}
S.~Gangapurwala, M.~Geisert, R.~Orsolino, M.~Fallon, and I.~Havoutis, ``Rloc: Terrain-aware legged locomotion using reinforcement learning and optimal control,'' \emph{IEEE Transactions on Robotics}, vol.~38, no.~5, pp. 2908--2927, 2022.

\bibitem{grcic2021dense}
M.~Grcic, P.~Bevandic, Z.~Kalafatic, and S.~{\v{S}}egvic, ``Dense anomaly detection by robust learning on synthetic negative data,'' \emph{arXiv preprint arXiv:2112.12833}, vol.~2, no.~5, 2021.

\bibitem{li2017neural}
W.~Li, N.~S. Szczecinski, and R.~D. Quinn, ``A neural network with central pattern generators entrained by sensory feedback controls walking of a bipedal model,'' \emph{Bioinspiration \& biomimetics}, vol.~12, no.~6, p. 065002, 2017.

\bibitem{goldsmith2020neurodynamic}
C.~Goldsmith, N.~S. Szczecinski, and R.~D. Quinn, ``Neurodynamic modeling of the fruit fly drosophila melanogaster,'' \emph{Bioinspiration \& biomimetics}, vol.~15, no.~6, p. 065003, 2020.

\bibitem{ijspeert2007swimming}
A.~J. Ijspeert, A.~Crespi, D.~Ryczko, and J.-M. Cabelguen, ``From swimming to walking with a salamander robot driven by a spinal cord model,'' \emph{science}, vol. 315, no. 5817, pp. 1416--1420, 2007.

\bibitem{zaier2008cpg}
R.~Zaier and S.~Kanda, ``Adaptive locomotion controller and reflex system for humanoid robots,'' in \emph{2008 IEEE/RSJ International Conference on Intelligent Robots and Systems}, 2008, pp. 2492--2497.

\bibitem{reynolds1987flocks}
C.~W. Reynolds, ``Flocks, herds and schools: A distributed behavioral model,'' in \emph{Proceedings of the 14th annual conference on Computer graphics and interactive techniques}, 1987, pp. 25--34.

\bibitem{rubenstein2014programmable}
M.~Rubenstein, A.~Cornejo, and R.~Nagpal, ``Programmable self-assembly in a thousand-robot swarm,'' \emph{Science}, vol. 345, no. 6198, pp. 795--799, 2014.

\bibitem{baldassarre2007sbots}
G.~Baldassarre, V.~Trianni, M.~Bonani, F.~Mondada, M.~Dorigo, and S.~Nolfi, ``Self-organized coordinated motion in groups of physically connected robots,'' \emph{IEEE Transactions on Systems, Man, and Cybernetics, Part B (Cybernetics)}, vol.~37, no.~1, pp. 224--239, 2007.

\bibitem{ozkan2021self}
Y.~Ozkan-Aydin and D.~I. Goldman, ``Self-reconfigurable multilegged robot swarms collectively accomplish challenging terradynamic tasks,'' \emph{Science Robotics}, vol.~6, no.~56, p. eabf1628, 2021.

\bibitem{fitzhugh1955mathematical}
R.~FitzHugh, ``Mathematical models of threshold phenomena in the nerve membrane,'' \emph{The bulletin of mathematical biophysics}, vol.~17, pp. 257--278, 1955.

\bibitem{cheng2015design}
C.~Cheng, P.~R. McGonigal, J.~F. Stoddart, and R.~D. Astumian, ``Design and synthesis of nonequilibrium systems,'' \emph{ACS nano}, vol.~9, no.~9, pp. 8672--8688, 2015.

\bibitem{chen2014dynamic}
Z.~Chen, T.~Chen, X.~Sun, and B.~J. Hinds, ``Dynamic electrochemical membranes for continuous affinity protein separation,'' \emph{Advanced functional materials}, vol.~24, no.~27, pp. 4317--4323, 2014.

\bibitem{bacstuug2012molecular}
T.~Ba{\c{s}}tu{\u{g}} and S.~Kuyucak, ``Molecular dynamics simulations of membrane proteins,'' \emph{Biophysical reviews}, vol.~4, pp. 271--282, 2012.

\bibitem{veltkamp2001shape}
R.~C. Veltkamp, ``Shape matching: Similarity measures and algorithms,'' in \emph{Proceedings International Conference on Shape Modeling and Applications}.\hskip 1em plus 0.5em minus 0.4em\relax IEEE, 2001, pp. 188--197.

\end{thebibliography}

\end{document}